\documentclass[runningheads]{llncs}
\usepackage[T1]{fontenc}
\usepackage{graphicx}
\usepackage{booktabs}
\usepackage{multirow}
\usepackage{setspace}
\usepackage[a4paper, left = 3.25cm, right = 3.25cm, top=3.25cm, bottom=3.25cm]{geometry}  
 
\begin{document}
\title{DAPE:Dynamic Non-uniform Alignment and Progressive Detail Enhancement Techniques for Improving the Performance of Efficient Visual Language Models}
%
%
\author{Mengyuan Tian\inst{1, 2} \and
Qiyan Zhao\inst{1, 2} \and
Yanan He*\inst{1} \and
Da-Han Wang\inst{2}}
\authorrunning{F. Author et al.}
%
\institute{
The Wang Yanan Institute for Studies
in Economics, Xiamen University, Xiamen 361005, China
\and Fujian Key Laboratory of Pattern Recognition and Image Understanding, School of Computer and Information Engineering, Xiamen University
of Technology, Xiamen 361024, China
 \\
}
\maketitle              
\begin{abstract}
In recent years, pre-trained visual-linguistic models have demonstrated tremendous potential, becoming a crucial foundational framework for numerous downstream tasks. However, the information density between text and images is not uniformly distributed. Existing methods often overlook the inherent and dynamic differences in information density and semantic scope between text tags and image blocks. These common uniform alignment strategies result in coarse-grained cross-modal interactions and loss of fine semantic details. Moreover, pursuing finer alignment typically requires substantial computational overhead, limiting practical model deployment. To address this challenge, this paper proposes a novel framework for dynamic cross-modal alignment with continuous detail introduction. First, we design a dynamically adaptive cross-modal matching mechanism that uses a learnable matching function to dynamically assign varying numbers and sizes of image tags to text tags of the same size but different information density, enabling more precise attention interaction. Second, we develop a continuous detail introduction module to progressively incorporate high-resolution visual feature enhancement into the alignment process. Extensive experiments across multiple benchmarks demonstrate significant improvements in the accuracy of various downstream tasks while reducing computational overhead.

\end{abstract}
\section{Introduction}
In recent years, pre-trained visual language models~\cite{zhao2025mca,zhao2026context,zhang2026hallucination,yip2026c,zhao2025textmamba,yip2026sope,zhang2024edge,shen2024multi,shen2025optimized,zhao2025few,zhang1,zhang2,zhang3,zhang4,zhang5,zhang6,zhang14,xu2026reasoning,ding2024text,zeng2025improved,zhao2024lightweight} based on large-scale data have achieved remarkable success in jointly modeling information from heterogeneous sources. By guiding pre-trained models to focus on information relevant to downstream tasks, this approach has yielded significant results\cite{21,22,23}. In complex, fine-grained scenarios, mainstream downstream tasks demand exceptional alignment capabilities between image attributes and textual semantics. The models must not only comprehensively understand multimodal information but also accurately model the consistency between image context and text.

Consider a typical image-text pair as an example: On the text side, in the sentence "In the forest, a yellow butterfly is fluttering above the head of a black-and-white dog," the information density of core semantic elements like "butterfly" and "dog" far exceeds that of functional words or modifiers such as "one,"  "of," and "is." Moreover, the visual area corresponding to "dog" is significantly larger than that of "butterfly." On the image side, the semantic richness of the image segments representing the dog's location, the butterfly's position, and their dynamic relationship far surpasses that of the expansive forest background. If uniform segmentation and equal-weight alignment are employed, the model consumes excessive computational resources on low-information-density background elements and functional words, neglecting the truly critical fine-grained semantic regions, resulting in blurred or misaligned alignment. In complex scenarios involving fine targets—such as the description "red packaged chips on the second row of the third shelf on the left" —precise target localization becomes essential. Although the target itself occupies only a small image area, its semantic importance is paramount; meanwhile, surrounding similar shelves and products serve as low-information-density background interference. Simply increasing image resolution and token count would lead to unnecessary computational overhead. Information is thus unevenly distributed across both visual regions and semantic dimensions. Furthermore, the image features extracted by modern visual encoders typically constitute a high-dimensional tensor, where different channels correspond to distinct semantic attributes—some channels may be sensitive to color, others respond to texture patterns, while additional channels focus on shape contours or object components. When text descriptions contain specific attribute terms such as "red,"  "striped," or "round," the ideal alignment should establish precise correspondence between these textual elements and the channels in the image features that encode those attributes, rather than merely locating them within the spatial region containing the object.

However, we have identified a fundamental limitation in current mainstream multimodal approaches: they typically construct uniform, static interaction templates for all input instances. This approach overlooks the non-uniform distribution of information density between textual and visual data. In text, core vocabulary such as nouns and verbs carry significantly higher information density than functional words like prepositions and articles; in images, regions containing key objects contain richer semantic information than flat backgrounds. Existing methods either use a fixed number of image labels to process uniformly sized text labels or treat all image regions equally. This coarse-grained alignment inevitably leads to ambiguity and loss of semantic information, becoming a bottleneck for further performance improvement. Conversely, pursuing finer alignment (e.g., using higher-resolution input images) directly results in exponential growth in computational complexity, making the model's speed and efficiency in practical applications unsatisfactory.

To address this challenge, we propose a novel Dynamic Alignment and Progressive Detail Enhancement Framework (DAPE), designed to fundamentally resolve the non-uniform information density distribution in cross-modal alignment while balancing computational efficiency. The framework consists of two core modules:  1. A dynamic adaptive matching mechanism comprising Channel-Level Alignment (CWA) and Non-Uniform Fine-Granularity Alignment (NFA). CWA decomposes image features into semantic segments along the channel dimension to achieve precise matching with text attributes, while NFA dynamically allocates multi-scale image blocks for differentiated interaction based on variations in text label information density—collectively overcoming the issue of ambiguous semantic matching caused by traditional uniform alignment.  2. A Progressive High-Frequency Detail Injection (PHI) module that employs learnable tokens to progressively "recall" high-frequency details such as edges and textures lost during downsampling from original high-resolution images, enabling refined feature enhancement without significantly increasing computational overhead.  The synergistic interaction of these modules enables the model to achieve precise semantic focus and detail completion while maintaining efficient inference, significantly improving the accuracy and robustness of cross-modal understanding. Experimental results demonstrate that DAPE achieves state-of-the-art performance across multiple benchmark tests.

Our main contributions are summarized as follows: 1)We propose a novel dynamic alignment and progressive detail enhancement framework named DAPE, which addresses the limitations of traditional uniform and static alignment strategies—such as the dilution of critical semantic information, resulting in ambiguity and detail loss, as well as restricted model performance ceilings. 2)We propose a dynamic adaptive matching mechanism that integrates CWA and NFA. CWA decomposes image features into distinct semantic segments along the channel dimension, enabling precise response to text attributes; NFA dynamically allocates multi-scale image blocks based on the information density of text tags, facilitating more refined interactions in key semantic regions. 3)We introduce the PHDIM module, which progressively recovers edge and texture details lost during downsampling from original high-resolution images using learnable tokens, thereby enhancing feature detail representation without significantly increasing computational overhead.
The proposed DEPA architecture demonstrates state-of-the-art performance across six mainstream datasets without significantly compromising efficiency.

\section{Related works }

With the rapid advancement of multimodal models and visual encoders, significant progress has been made in various visual-language (VL) tasks. In recent years, numerous open-source multimodal visual language models have emerged. Pre-trained models align visual and textual representations into a shared semantic space, with some demonstrating exceptional performance in specific domains. For instance, models like CLIP \cite{1} and ALIGN employ contrastive learning on massive text-image pairs to develop transferable representations, showcasing strong zero-shot and few-shot capabilities across downstream tasks. Subsequent work such as ALBEF and BLIP further integrate cross-modal fusion mechanisms to enhance inter-modal interactions. These models typically adopt a uniform alignment strategy, assuming all tokens are equally important, thereby overlooking the inherent non-uniformity in cross-modal information density.

Cross-modal alignment remains a fundamental challenge in visual language modeling. Early approaches predominantly employed global or region-level matching to calculate similarity between images and text. These methods typically treated all image regions or text tokens equally, resulting in coarse-grained interactions. Preserving high-frequency details is crucial for fine-grained visual understanding. Traditional methods often relied on multi-scale feature fusion or high-resolution branch networks to retain details, which effectively enhanced detail capture capabilities \cite{2,3,4,5,6,7,8}. While attempts to achieve finer alignment through operations on visual tokens at the image block level have been made, they still face high computational costs when processing high-resolution inputs.

On improving model efficiency,\cite{16} propose contrastive distillation for visual representation learning. In addition, remarkable advances have been made in knowledge distillation for language model compression (i.e. , BERT \cite{10}), and these works show that mimicking the distribution of self-attention and intermediate representations of transformer blocks increases performances \cite{13,12,15,18}, for downstream tasks. In particular, in the transformer-based language model distillation, DistillBERT \cite{13} proposes to train the small BERT by mimicking the Teacher's output probability of masked language prediction and the embedding features. TinyBERT \cite{12} and MobileBERT \cite{15} leverage the layer-wise attention distributions for distillation with MSE function. \cite{17} suggests distilling on the last transformer layer and bringing extra flexibility for training. \cite{11,14},  also use the contrastive distillation in transformer based language model compression. \cite{9,14} propose using a sample queue to store history embeddings and show that contrasting with more negative samples is beneficial for knowledge distillation.These solutions are unrelated to the research focus of this paper. We enhance model efficiency by exploring network architecture design.

While existing methods have made substantial progress in visual language pre-training and alignment, they often overlook the non-uniform distribution of information across modalities and the trade-off between detail preservation and computational efficiency. Our proposed DAPE framework addresses these limitations by introducing a dynamic non-uniform alignment mechanism and a progressive detail injection strategy, thereby achieving precise cross-modal matching and efficient inference.

\section{Method}

Specifically, given an input image, the visual features are extracted using an image encoder, denoted as $ M \in {R}^{h \times w \times d}$ , where $h \times w$ represents the resolution and $d$ is the feature dimension. The text features $ {R}^{l \times d} $ are extracted using a text encoder.

\subsection{Operation of DAPE in the backbone section}
\subsubsection{Coarse Alignment of Visual and Text Features}

To reduce the computational load of the main component, the original image feature $M\in R ^{h \times w \times d} $ is downsampled to $ {M_0} \in R^{h/s \times w/s \times d}$. The image feature $ M_0$ and text feature  $T$ are then uniformly truncated into $I$ image tokens $m_{0_{i}}$ and $J$ text tokens $ t_j$ respectively.

The similarity between image token $m_{0_{i}}$ and text token $t_j$ is calculated using the cosine similarity $s({\cdot},{\cdot})$ , yielding an affinity matrix ${A_0}\in {R}^{I\times J}$ that measures the affinity between different image tokens and prompts, where $A_0[i, j] = s(m_{0_i}, t_j)$. Each element in matrix $A_0$ is binarized by setting values above a threshold $k_0$ to 1 and others to 0. Specifically, the binarized matrix ${A_0} \in {[0,1]}^{I \times J}$ is utilized in the cross-attention module.

\begin{equation}
Q_T=T * W_q
\end{equation}

\begin{equation}
K_T = T * W_k
\end{equation}

\begin{equation}
V_T = T * W_v
\end{equation}

Similarly, $Q_{M_0}$, $K_{M_0}$, and $V_{M_0}$ are obtained.

\begin{equation}
T_1=softmax\left ({\frac{Q_T{K_{M_0}^T}}{\sqrt{d}}}\right )\mathrm{·}{{A_0}^T}\mathrm{·}{V_{M_0}}
\end{equation}

\begin{equation}
M_1=softmax\left ({\frac{{Q_{M_0}}K_T^T}{\sqrt{d}}}\right )\mathrm{·}A_0\mathrm{·}V_T
\end{equation}

$M_1$ serves as the visual feature input for $M_0$ in the subsequent layer.

\begin{figure}[t]
\centering
\includegraphics[width=\textwidth]{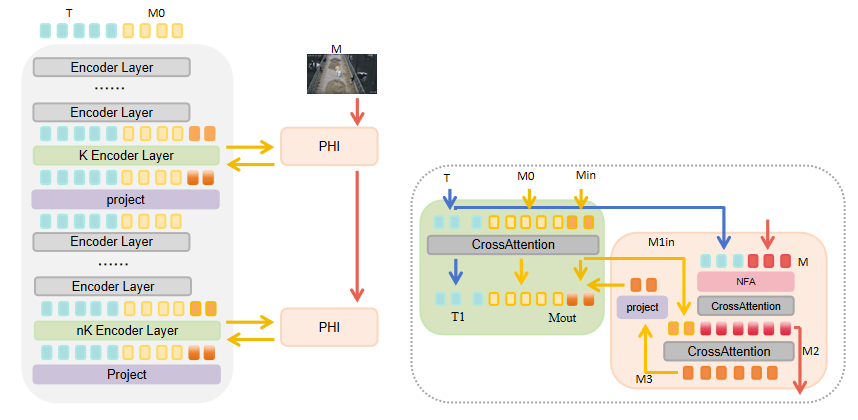}
\caption{Schematic diagram of the DAPE framework and dynamic non-uniform alignment process}\label{fig1}
\end{figure}

\subsubsection{Channel-Wise Text-Image Alignment}

Due to certain feature attributes in the text, the channel level representation in image features shows higher correspondence.
Replace the entire image $M_1$ to obtain $ C \in {R}^{d \times {w\times h}/{s^2}}$.The channel is uniformly divided into $L $segments, each of which is ${c_l} \in R^{{d/L} \times{{wh}/{s^2}}}$ .
Influenced by the SAP method, a two-layer MLP ($\cdot$) is directly used to predict the visual attention weight ${a} \in { R^d}$ : $a$ = Softmax(MLP(C)) = ($a_1$, $a_2$, ..., $a_l$). From each segment, the top$ k_1 $values are selected ( $b\in $ [ $l_1$,$l_2$,...,$l_{k_1}$] ), $l \in [1,L]$) and concatenated to form $B \in {R}^{L \times (w \times h) / s^2}$, where ${b_l} \in R^{wh/{s^2}}$, $ l\in [1, L]$.

\begin{figure}[t]
\centering
\includegraphics[width=\textwidth]{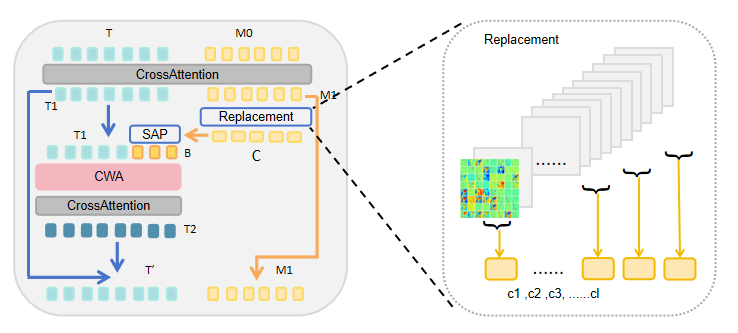}
\caption{Channel-level text-image alignment (CWA) module structure diagram}\label{fig1}
\end{figure}

Calculate the similarity matrix between the channel-direction image feature token bl and the text $b_l $token $t_{1_j}$,

\begin{equation}
s({b_l},{t_{1_j}})=\cos({b_l},{t_{1_j}})
=\frac{{b_l}\cdot{t_{1_j}}}{||{b_l}||\cdot||{t_{1_j}}||}
\end{equation}

Through the aforementioned process, we obtain an affinity matrix $A_c \in \{0,1\}^{L \times J}$ that measures the similarity between image feature tokens along the channel direction and different text prompts, where $A_c[l, j] = s(b_l, t_{1_j})$. Each element in matrix $A_c$ is binarized: values above the threshold kc are set to 1, while others are 0. Specifically, the binarized matrix $A_c \in \{0,1\}^{L \times J}$ is used as a query to interact with the image's channel feature $B$ through cross-attention, with $A_c$ serving as a mask matrix. As described in (4), this yields text features $T_2$. The resulting text features $T_1$ and $T_2$ are then combined to form $T'$, which serves as the input for the next layer's text feature input $T$.

\subsection{ Non-Uniform Fine-Grained Image Patch and Text Alignment Matching}

Different text token labels carry varying information densities, corresponding to image blocks with distinct feature characteristics. Using uniform matching parameters for all image and text tokens would result in feature extraction that lacks differentiation and is prone to bias. To address this, we innovatively introduce a coordinated prompt attention mechanism tailored to different information densities, enabling multi-sensory field interaction. This approach significantly enhances the pre-trained model's capability to extract rich and differentiated features from input images.

\begin{figure}[t]
\centering
\includegraphics[width=\textwidth]{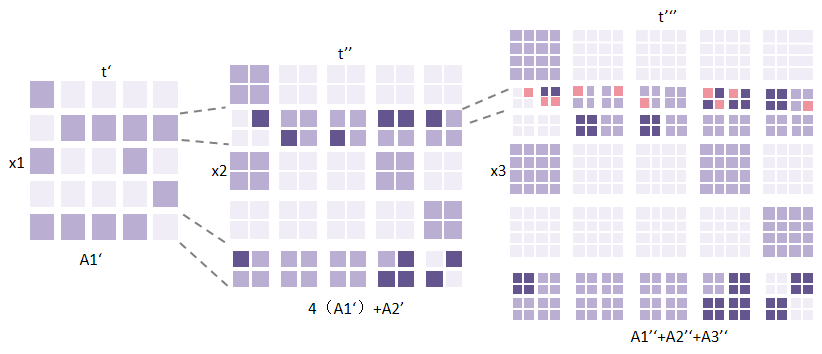}
\caption{Schematic diagram of Nonuniform Fine-grained Image Block and Text Alignment Matching (NFA) mechanism}\label{fig1}
\end{figure}

Fine-grained image features require higher-resolution images. For the input original image feature $M \in {R}^{h \times w \times c}$, it is divided into three parts along the channel: it is divided along the channel dimension into three parts. Each part, $X_1$, $X_2$, $X_3 \in {R}^{h \times w \times \mu_1 c} ,{R}^{h \times w \times \mu_2 c} , {R}^{h \times w \times \mu_3 c}$ (where $ {\mu_1} +{\mu_2}  + {\mu_3} = 1$ ). Local convolution operations with different kernel sizes are applied to obtain $X'_1$, $X'_2$, $X'_3 \in {R}^{h \times w \times \mu_1 c} ,{R}^{h \times w \times \mu_2 c} , {R}^{h \times w \times \mu_3 c}$ . These are then uniformly cropped into image tokens $x_1$, $x_2$, and $x_3$ in a 1:2:4 ratio. Simultaneously, the text feature T is uniformly cropped into text tokens $t$.

The affinity matrix ${A_1}\in R^{I×J}$ is computed for $x_1$ and $t'$ using the method described in 3.1.1, where ${A_1ij} = S({x_{1_i}}, {t'_ j})$. Each element in $A_1$ is binarized: elements above the threshold k are assigned$ \mu_1$, while others are set to 0, yielding ${A'_1}^{I\times J}\in [0, {\mu_1}]$ scaled by a factor of 16. Specifically, the binarized matrix ${A''_1}^{4I\times4J}\in [0, {\mu_1}]$ is obtained.

When a certain number of rows contain $\mu_1$ values exceeding a specified threshold, it indicates that the corresponding image tokens exhibit high information density. For the $x_2$ position, we extract the corresponding token, divide each $t'$equally into twice the original size, and calculate the similarity between these tokens to generate an affinity matrix ${A_2} \in R^{2I\times 2J}$ . For positions with high image information density, we compute$A_2^{{2I}\times{2J}} = S({x_{2_i}}, {t'_ j})$ , while setting all other positions to 0. We then binarize each element in $A_2$ by assigning $\mu_2$ to values above the threshold k and 0 to others, yielding ${A'_2} ^{2I\times2J}\in [0, {\mu_2}]$. This matrix is proportionally scaled to four times its original size. Specifically, the binarized matrix ${A"_2} ^{4I\times4J}\in [0, {\mu_2}]$ is obtained. Using the same method, we derive ${A"_3} ^{4I\times4J}\in [0, {\mu_3}]$.

\begin{equation}
A’={A''_1}+{A''_2}+{A''_3}
\end{equation}

\begin{equation}
M2 = softmax\left ({\frac{Q_MV_T^T}{\sqrt{d}}}\right )\mathrm{·}A'\mathrm{·}V_T
\end{equation}

\subsection{Progressive High-Frequency Detail Injection Mechanism}

As mentioned previously, pursuing extreme fine-grained alignment (as discussed in Sections 3.1.2 and 3.1.3) introduces non-negligible computational overhead. However, simply reducing the input image resolution to improve speed inevitably leads to the loss of high-frequency details crucial for recognition, such as object edges, textures, and smaller entities. This loss of detail directly impairs the model’s representational capacity, especially when dealing with small objects or complex scenes. To resolve this contradiction, we propose a Progressive High-Frequency Detail Injection Mechanism(PHI).The core idea is to selectively supplement the detail information contained in the original high-resolution image into the compressed features undergoing alignment, in a controllable and learnable manner, without significantly increasing the main computational burden.

The input image feature $M_0$ from the final layer of each K-level hierarchy is partially filled with learnable tokens $M_{in}$. Through a visual text feature rough alignment process, this generates text feature $T_1$ and image feature ${M_1}^+$. Here, $T_1$ serves as the input for the next layer's text feature, while the originally filled learnable token portion in ${M_1}^+$ becomes $M_{1_{in}}$, which is then fed into the progressive high-frequency detail injection module as a query. Simultaneously, the high-resolution image $M$ (enhanced via Fourier transform) undergoes a learnable linear projection layer to convert it into a set of detail tokens as keys and values. These tokens interact with the k-1 layer's text feature output $T'$ through a non-uniform fine-grained image-text alignment process to generate image feature $M_2$. This step specifically uses the original high-resolution image because non-uniform adaptive image-text alignment requires extracting more explicit detail features from higher-resolution image features.

In the final layer of every K-th block, the input image feature $M_0$ is padded with a set of learnable tokens $M_{in}$ . Together with the text feature input $T$, they undergo the coarse visual-text feature alignment process, producing the output text feature $T_1$ and the image feature ${M_1}^+$. Here, $T_1$ serves as the text feature input $T$ for the next layer, while the portion in ${M_1}^+$ that was originally the padded learnable tokens is extracted as $M_{in}$ and input into the PHI.

Simultaneously, the original high-resolution image M (classically augmented and transformed) is passed through a learnable linear projection layer to be converted into a set of detail tokens, serving as the key and value. These detail tokens, along with the text feature output $T$ from the (k-1)-th layer, undergo a non-uniform fine-grained image patch and text alignment interaction to obtain the image feature $M_2$. The reason for using the original high-resolution image here is that the non-uniform adaptive image patch and text alignment requires more explicit detailed features from higher-resolution image representations.

$M_{in}$ and $M_2$ then engage in a cross-attention interaction, yielding the further-aligned and detail-augmented image feature $M_3$. The purpose is to allow the current alignment feature $M_{in}$ to actively "query" and "recall" from the high-resolution, non-uniformly aligned global detail pool, thereby retrieving those critical detail pieces that might have been lost during the downsampling process but are essential for the current alignment task.

The detail-enhanced feature $M_2$ is then connected to $M_{in}$ via a residual connection, replacing the original $M_{in}$ position within ${M_1}^+$ to obtain the detail-enhanced image feature $M_{out}$, which serves as the input $ M_0$ for the (K+1)-th layer.

The interacted high-resolution image detail feature $M_2$ is preserved and will be used as the "original high-resolution image" for the next detail injection step (at the n(K+1)-th instance). This design allows detail information to be transmitted and evolved across the deep network layers, forming a continuous, progressive enhancement process that ensures the coherence of detail supplementation.

\section{Experiment}

\subsection{Datasets}

\textbf{RefCOCO}\cite{18} This is a series of datasets built on MS COCO images, specifically designed for the "referential expression understanding" task, which involves locating a target object in an image based on a natural language description. It acts as a key bridge connecting computer vision and natural language processing.

\textbf{CIFAR-100}\cite{19}A classic small-scale image classification dataset comprising 60,00032×32 pixel color images, categorized into 100 fine-grained classes. Each image belongs to a single category, commonly used for lightweight model training and algorithm benchmarking.

\textbf{GRefCOCO}\cite{20}It is a large-scale dataset for general referring expression segmentation. It not only includes traditional single-target samples but also innovatively introduces two types of complex samples: multi-target (one expression refers to multiple instances) and no-target (the described object does not exist in the image). With a total of approximately 278,000 expressions, the dataset aims to advance models' ability to handle diverse and ambiguous instructions in real-world scenarios.

\textbf{T-Cows}The video data was collected in 2025 from a dairy farm located approximately 80 kilometers northeast of Baotou City, Inner Mongolia Autonomous Region, involving 52 lactating cows. The dataset comprises 3,84045-minute 1080p video clips. Using keyframe extraction software, the videos were converted into keyframe sequences. After manual screening that considered factors such as lighting variations, cow occlusion, object overlap, and positive-negative sample balance, 2,410 images were ultimately selected to form this dataset.

To effectively train and validate our proposed dynamic fine-grained multimodal alignment model, we meticulously constructed a large-scale, high-quality "visual-text" dairy cow behavior recognition dataset. Each key frame or short sequence was accompanied by detailed, contextually clear descriptions, such as: "A cow is burying its head deep in the feeding trough, mouth and nose continuously touching the feed, exhibiting typical feeding posture. In the distance, another cow stands on all fours, away from the trough, in a wandering state."

\subsection{Implementation Details}
All experiments were conducted on two NVIDIA RTX 4090 GPUs. The batch size for all tasks was set to 8.In the dynamic alignment module, the binarization threshold is set to \(k_0 = 0.5\). For CWA, the channel dimension is divided into \(L = 8\) segments, with the first \(k_1 = 4\) channels selected from each segment. In NFA, the partition ratio is configured as \(\mu_1: \mu_2: \mu_3 = 1:2:4\), with a threshold of \(k = 0.6\). Finally, the PHI mechanism is applied every 4 layers to incorporate high-resolution details.

\subsection{Comparison with State-of-the-arts}

To comprehensively evaluate the effectiveness of the DAPE method, we conducted comparative experiments with existing advanced methods across multiple visual language tasks. The experiments covered four major tasks: Visaul Grounding, image classification, image-text retrieval,  and Generalization refers to representational understanding (GREC), and were validated on datasets including RefCOCO, CIFAR-100, GRefCOCO, and T-Cows.

\begin{table}[htbp]
\centering
\caption{Performance comparison of Visaul Grounding models integrated with DAPE on RefCOCO and T-Cows datasets}
\resizebox{\linewidth}{!}{
\begin{tabular}{cccccccccccc}
\toprule
\multirow{2}{*}{\textbf{Method}} &  \multicolumn{2}{c}{\textbf{RefCOCO}} & &&\multicolumn{2}{c}{\textbf{T-Cows}} & \\
 &  \textbf{mAP50(\%)} &\textbf{mAP75(\%)} &\textbf{FPS} &&  \textbf{mAP50(\%)} &\textbf{mAP75(\%)} &&\textbf{FPS}\\
\midrule
UNINEXT &  82.8&63.4 &  13.0&&96.8 & 81.8 && 15.2 \\
+DAPE &  84.1 &66.0&13.5 & &97.2  & 84.6& &15.8  \\
\hline
RealGIN&  82.1 &65.0 & 9.2& &93.9 &80.0&&11.1\\
+DAPE &  83.3 &65.8&9.3 &&95.2 &84.1&&11.9\\
\bottomrule
\end{tabular}
}
\label{tab:zero_shot_evaluation}
\end{table} 

In the Visaul Grounding task (Table 1), we evaluated two models—UNINEXT and RealGIN—on the COCO and T-Cows datasets. The results demonstrated that the integration of the DAPE method significantly improved the detection accuracy of both models. Specifically, the UNINEXT model achieved 1.3\% and 2.6\% improvements in mAP50 and mAP75 on the RefCOCO dataset, and 0.4\% and 2.8\% on the T-Cows dataset, respectively. The RealGIN model showed 1.2\% and 1.8\% improvements in mAP50 and mAP75 on the RefCOCO dataset, and 1.3\% and 4.1\% on the T-Cows dataset.

\begin{table}[htbp]
\centering
\caption{Classification accuracy and inference speed of vision-language models enhanced with DAPE on CIFAR‑100}
\resizebox{0.5\linewidth}{!}{ 
\begin{tabular}{cccccccc}
\toprule
\multirow{2}{*}{\textbf{Method}}   & & &\multicolumn{2}{c}{\textbf{CIFAR-100}} & && \\
 &  \textbf{Top1} & &\textbf{Top5} & &\textbf{FPS} 
 \\
\midrule
EVA-CLIP &  71.8& & 87.4  & & 16.8    \\
+DAPE &  72.6 &+(0.6)&88.5  &+(1.1) & 17.0  \\
\hline
VFPT &  62.1 && 79.3  &&   10.9\\
+DAPE &  64.7 &+(2.6)&  82.2 &+(2.8)&9.7 \\
\hline
VPT &  65.5 && 73.3 &&  14.2 \\
+DAPE &  67.0 &+(1.5)&  75.5 &+(2.2)& 14.5 \\
\hline
DAMVP & 73.1  &&  87.9 &&   18.0\\
+DAPE &  76.5 &+(3.4)& 88.8  &+(0.9) &19.1  \\
\bottomrule
\end{tabular}
}
\label{tab:zero_shot_evaluation}
\end{table}

In the image classification task (Table 2), we evaluated four models—EVA-CLIP, VFPT, VPT, and DAMVP—on the CIFAR-100 dataset. The results demonstrated that DAPE consistently improved the classification accuracy of all models. For instance, on CIFAR-100, DAMVP with DAPE achieved 3.4\% and 0.9\% improvements in Top1 and Top5 accuracy, respectively. Notably, the inference speed of all models remained stable, with some (e.g., VPT) even achieving higher FPS on CIFAR-100. This indicates that DAPE enhances feature discrimination without significantly increasing computational load.

In the text-image retrieval task (Table 3), the LAPS and CUSA models demonstrated improved performance on both RefCOCO and T-Cows datasets after integration with DAPE. For instance, the CUSA model achieved a 2.6\% improvement in text retrieval R@1 and a 2.8\% improvement in image retrieval R@1 on RefCOCO, while on T-Cows, it showed a 3.1\% improvement in text retrieval R@1 and a 3.5\% improvement in image retrieval R@1. These results demonstrate that DAPE significantly enhances the accuracy and robustness of cross-modal retrieval through its dynamic alignment and detail enhancement mechanisms.

The experimental results demonstrate that the DAPE method significantly improves model performance across various visual language tasks, particularly in enhancing feature discriminability and cross-modal alignment quality while maintaining high inference efficiency. This validates its practicality and generalization capability as a plug-and-play enhancement module.

\begin{table}[htbp]
\centering
\caption{Image‑text retrieval performance of models with and without DAPE on RefCOCO and T-Cows}
\resizebox{\linewidth}{!}{
\label{tab:coco_results}
\begin{tabular}{lccccccccccccccc}
\toprule
& &\multicolumn{3}{c}{\textbf{RefCOCO}}&&&&\multicolumn{3}{c}{\textbf{T-Cows}}\\
 Method & \multicolumn{3}{c}{\textbf{Text Retrieval}} & \multicolumn{3}{c}{\textbf{Image Retrieval}} &&\multicolumn{3}{c}{\textbf{Text Retrieval}} & \multicolumn{3}{c}{\textbf{Image Retrieval}} \\
 & R@1 && R@5 & R@1& & R@5 &rSum&  R@1 & &R@5 & R@1 && R@5 &rSum&   \\
\midrule
LAPS & 58.8 && 84.3 &  44.5 & &74.3 &261.9& 69.2 & &87.8 & 75.9 && 85.4&318.3 \\
+DAPE & 61.6 && 85.7 &  45.1 && 76.6 & 269.0& 71.0 & &88.3 &   76.1 && 88.8 & 324.2 \\
\hline
CUSA & 61.2 && 86.8 &  56.2 && 75.9 & 280.1& 71.2 & &85.0& 75.6 && 81.2& 313.0 \\
+DAPE & 63.8 && 87.2  &  59.0&& 78.6 & 288.6& 74.3 & &86.2&  79.1 & &82.4 & 322.0\\
\bottomrule
\end{tabular}
}
\end{table}

Table 4 presents the performance comparison of different models with and without DAPE on the GREC task using the GRefCOCO dataset. The evaluation metrics are Prec@(F1@0.5) and N-acc. The results show that both MDETR and UNINEXT achieve improvements after adding DAPE: MDETR gains 2.8 and 1.4 percentage points, respectively, while UNINEXT gains 2.6 and 0.4 percentage points, indicating that DAPE effectively enhances the models’ referring expression comprehension capability.

\begin{table}[htbp]
\centering
\caption{Generalization refers to representational understanding (GREC) performance of models with and without DAPE on GRefCOCO}
\resizebox{0.5\linewidth}{!}{
\label{tab:coco_results}
\begin{tabular}{lcccccccccccccc}
\toprule
 Method & \multicolumn{3}{c}{\textbf{GRefCOCO}} &  && &  \\
 & Prec@(F1@0.5) && N-acc.   \\
\midrule
MDETR & 58.8 && 84.3  \\
+DAPE & 61.6 && 85.7 \\
\hline
UNINEXT & 61.2 && 86.8\\
+DAPE & 63.8 && 87.2 \\
\bottomrule
\end{tabular}
}
\end{table}

\subsection{Ablation}

\subsubsection{Effectiveness of Channel-Wise Text-Image Alignment (CWA)}

The core innovation of the CWA module lies in decomposing image features into semantic fragments along the channel dimension and adaptively aligning them with text features, enabling fine-grained matching at the attribute or concept level. For example, when describing "a red high-speed train," CWA enhances the correlation between "red" and color channel features, and between "train" and shape channel features, outperforming global pooling or uniform interaction. To validate its effectiveness, we integrated CWA into two vision-language foundation models, UNINEXT and VPT, for visual grounding and image recognition tasks, using raw high-resolution images as input. As shown in Table 6, with CWA integrated, UNINEXT achieves a 0.3\% improvement in mAP50 and 0.9\% in mAP75 on RefCOCO visual grounding tasks, and a 0.7\% improvement in mAP75 on T-Cows visual grounding tasks. As shown in Table 5, on the CIFR-100 image classification task, Top-1 and Top-5 accuracy improve by 0.3\% and 0.4\% respectively over the baseline.

\begin{table}[htbp]
\centering
\caption{Ablation study of individual components in DAPE on  image classification tasks}
\resizebox{0.4\linewidth}{!}{
\begin{tabular}{lccccccccccccccc}
\toprule
\multirow{2}{*}{\textbf{Method}}\\
   &\multicolumn{2}{c}{\textbf{CIFAR-100}} & & \\
 &  \textbf{Top1} & \textbf{Top5} & \textbf{FPS} &\\
 \hline
 VPT &  65.5 & 73.3 &  14.2 \\
+CWA &  65.8 &72.7 & 10.4 \\
+NFA &  66.6& 74.9& 8.9 \\
+PHI &  63.8& 69.7 & 15.8 \\
+DAPE &  67.0 &  75.5 & 14.5\\
\bottomrule
\end{tabular}
}
\label{tab:zero_shot_evaluation}
\end{table}

\begin{table}[htbp]
\centering
\caption{Ablation study of individual components in DAPE on Visaul Grounding tasks}
\resizebox{\linewidth}{!}{
\begin{tabular}{lccccccccccccccc}
\toprule
\multirow{2}{*}{\textbf{Method}}\\
& \multicolumn{2}{c}{\textbf{RefCOCO}}&&\multicolumn{2}{c}{\textbf{T-Cows}} & \\
 &  \textbf{mAP50(\%)} &\textbf{mAP75(\%)} &\textbf{FPS} &\textbf{mAP50(\%)} &\textbf{mAP75(\%)}&\textbf{FPS}& &\\
\midrule
UNINEXT &  82.8&63.4 &  13.0&96.8 & 81.8 & 15.2 \\
+CWA &  83.1 &64.3& 12.3  &96.6& 82.5&14.2 \\
+NFA &  85.0 &65.3 &9.1 &97.0&84.8&10.5\\
+PHI &  82.4 &62.8 & 17.1 & 92.8&82.0&17.9\\
+DAPE &  84.1 &66.0&13.5  &97.2  & 84.6 &15.8  \\
\bottomrule
\end{tabular}
}
\label{tab:zero_shot_evaluation}
\end{table}

\subsubsection{Non-Uniform Fine-Grained Image Patch and Text Alignment Matching (NFA)}

The NFA module addresses the limitation of static interaction templates. For “a butterfly perched on a flower,” the baseline uniformly interacts with all words (e.g., “a,” “on,” “flower”), diluting attention to key entities like “butterfly” and “flower.” NFA dynamically allocates larger receptive fields to information-dense concepts (e.g., “butterfly”) for fine-grained interaction, while sparing resources for sparse words (“on”), improving semantic focus and feature discrimination.

We added NFA only to the encoder layers of vision-language models (no PHI module), directly into the backbone with original high-resolution images as input. As shown in Table 6, on RefCOCO visual grounding, NFA improves UNINEXT by 2.2\% in mAP50 and 1.9\% in mAP75; on T-Cows, mAP50 rises 0.2\% and mAP75 by 3.0\%. For image classification (Table 5, CIFR-100), Top-1 and Top-5 accuracy increase by 1.1\% and 1.6\% over baseline. However, without progressive injection and with direct backbone integration, model efficiency declines.

\subsubsection{Effectiveness of Progressive High-Frequency Detail Injection Mechanism (PHI)}

To evaluate the PHI module alone, we added it to the baseline (“baseline + PHI”). PHI addresses high-frequency detail loss from downsampling by enabling initial cross-attention between original high-frequency image labels and text labels. On RefCOCO visual grounding, PHI improved UNINEXT’s FPS by 4.1\%; on T-Cows, FPS increased by 2.7\%; on CIFR-100 image classification, FPS improved by 1.6\%. Results show PHI boosts speed without significant accuracy loss.

\subsubsection{Analysis of the Balance between Computing Efficiency and Performance}
As shown in the ablation results in Tables 5 and 6, introducing the NFA module alone significantly increases computational overhead, mainly because it establishes dynamic, multi-granularity attention interactions in information-dense regions, requiring more visual tokens and complex similarity calculations. Introducing the PHI module alone can improve inference speed (e.g., FPS on RefCOCO increases from 13.0 to 17.1), as PHI selectively extracts and injects details from the original high-resolution image via a cross-attention mechanism, avoiding global processing and reducing redundant computation. The DAPE framework synergistically integrates NFA, PHI, and CWA modules to achieve an optimal balance between efficiency and performance: NFA enhances alignment precision, PHI compensates for performance loss from feature degradation while optimizing computational workflows, and CWA provides channel-level auxiliary alignment. As shown in Table 5, the full DAPE framework demonstrates strong comprehensive performance across multiple tasks: in the T-Cows task, TCM+DAPE achieves significant improvements in mAP50 and mAP75 while maintaining an FPS of 15.8, comparable to the baseline (15.2); in the VPT classification task, Top-1/Top-5 accuracy on CIFAR-100 is substantially improved, with FPS stable or even slightly higher than the baseline. Therefore, DAPE does not simply stack modules but achieves fine-grained alignment and detail enhancement of key semantics through dynamic resource allocation and progressive detail fusion, while maintaining high inference efficiency, thus achieving an optimal balance between computational cost and performance gain in practical applications.

\section{Conclusion}

This study introduces the DAPE framework, which effectively addresses the imbalance between information density distribution and detail retention with computational efficiency in visual language models through dynamic non-uniform alignment mechanisms and progressive detail enhancement techniques. The method adaptively allocates cross-modal attention resources while selectively injecting high-frequency visual details, achieving significant performance improvements across multiple benchmark tasks without substantially increasing computational overhead. Future work may explore extending this framework to video-language tasks, lightweight deployment, and integration with emerging multimodal architectures, further advancing the development of efficient and precise cross-modal understanding technologies.

\section{Acknowledgement}
This work is supported by the Major Science and Technology Plan Project on the Future Industry Fields of Xiamen City (No. 3502Z20241027), and National Natural Science Foundation of China (No. 62576301).

%
%
%
\bibliographystyle{splncs04}
\bibliography{mybibliography}
\end{document}